\documentclass[11pt, a4paper, twocolumn, copyright, gdm]{google}

\usepackage[authoryear, sort&compress, round]{natbib}
\usepackage{tabularx}   
\usepackage{ragged2e}

\uselogo{} 

\title{Architecting Trust in Artificial Epistemic Agents}

\correspondingauthor{nahemamarchal@google.com}

\author[1]{Nahema Marchal}
\author[1]{Stephanie Chan}
\author[1]{Matija Franklin}
\author[1]{Manon Revel}
\author[2]{Geoff Keeling}
\author[2]{Roberta Fischli}
\author[1]{Bilva Chandra}
\author[1]{Iason Gabriel}

\affil[1]{Google DeepMind}
\affil[2]{Google Research}

\begin{abstract}

Large language models increasingly function as epistemic agents—entities that can (1) autonomously pursue epistemic goals and (2) actively shape our shared knowledge environment. They curate the information we receive, often supplanting traditional search-based methods, and are frequently used to generate both personal and deeply specialized advice. How they perform these functions, including whether they are reliable and properly calibrated to both individual and collective epistemic norms, is therefore highly consequential for the choices we make. We argue that the potential impact of epistemic AI agents on practices of knowledge creation, curation and synthesis, particularly in the context of complex multi-agent interactions, creates new informational interdependencies that necessitate a fundamental shift in evaluation and governance of AI. While a well-calibrated ecosystem can augment human judgment and collective decision-making, poorly aligned agents risk causing cognitive deskilling and "epistemic drift," making the calibration of these models to human norms a high-stakes necessity. To ensure a beneficial human-AI knowledge ecosystem, we propose a framework centered on building and cultivating the trustworthiness of epistemic AI agents; aligning AI these agents with human epistemic goals; and reinforcing the surrounding socio-epistemic infrastructure. In this context, trustworthy AI agents must demonstrate epistemic competence, robust falsifiability, and epistemically virtuous behaviors, supported by technical provenance systems and "knowledge sanctuaries" designed to protect human resilience. This normative roadmap provides a path toward ensuring that future AI systems act as reliable partners in a robust and inclusive knowledge ecosystem.

\end{abstract}

\begin{document}

\maketitle

\section{Introduction}
Multimodal large language models\footnote{LLMs are advanced AI models with large context windows, trained on large-scale datasets (of text, images, video, and code) then fine-tuned through supervised and reinforcement learning techniques, allowing them to understand language patterns and generate human-like outputs across modalities. A key feature of these models is their ability to store a vast amount of information about the world within their parameters (referred to as parametric knowledge) during pre-training, which can then be extracted through question-answering.} (LLMs) are profoundly reshaping our epistemic practices— how we acquire, produce and interpret information. Recent estimates suggest that approximately 9\% of US news articles published are either partially or fully generated by artificial intelligence (AI) \citep{russell2025}, while entire encyclopedias and knowledge bases are now being created with these models (e.g., \citep{Chris_2025}). The past years have also seen an unprecedented surge in the use of LLMs for knowledge-related tasks \citep{chatterji2025, appel2025, spencer2025, wasik2025, Li_2025}. Their interactive and conversational nature has positioned them as primary tools for information-seeking, rivaling traditional search engines for tasks that require nuanced comprehension and synthesis \citep{Greussing, sun2024trustingsearchunravelinghuman, caramancion2024largelanguagemodelsvs}. Globally, millions now turn to these systems as companions, thought partners and guides to help them interpret and navigate the world around them. Crucially, emerging research shows that these models are starting to impact what we know and how we act: from our political beliefs and preferences \citep{potter2024hiddenpersuadersllmspolitical, schoenegger2025largelanguagemodelspersuasive, hackenburg2025leverspoliticalpersuasionconversational, costello2024durably} to our vocabulary choices \citep{yakura2025empiricalevidencelargelanguage}, and our cognitive abilities \citep{lee2025impact, stadler2024cognitive}.

These AI systems are also becoming increasingly agentic and autonomous, capable of executing multi-step tasks, browsing the web, and acting on behalf of users with limited input \citep{kasirzadeh2025characterizingaiagentsalignment, lewis2020retrieval, huang2025deepresearchagentssystematic}. Furthermore, they are increasingly entrusted with knowledge production, validation and curation itself— a trend visible in the development of AI tutors \citep{jurenka2025responsibledevelopmentgenerativeai}, peer-reviewers \citep{perlis2025artificial} and scientists \citep{lu2024aiscientistfullyautomated, gottweis2025aicoscientist}. As the human-AI relationship shifts toward interaction with more autonomous, adaptive, and integrated agents, fundamental questions arise: When and on what basis should we trust the information provided by these agents? Under what conditions can they be trusted to participate in knowledge creation? And what processes need to be put in place, including at the collective level, to ensure that they reliably augment, rather than degrade, our information and knowledge ecosystems?

These questions carry critical real-world implications, especially given the complex web of interdependencies created by multi-agent ecosystems. Because no single individual has the temporal, cognitive or material capacity to master every subject, we are fundamentally epistemically dependent on the testimony and expertise of others—from scientists to journalists— to navigate the world \citep{hardwig1985epistemic}. In a world in which we may increasingly rely AI agents (which, in turn, interface with other AI agents) to mediate this knowledge and guide decision-making, understanding what justifies epistemic trust in these systems becomes paramount. Only with this understanding in place, can we establish clear desiderata and practical guidelines to evaluate whether these conditions are met, and whether different kinds of agent deployment support human autonomy and the integrity of our shared knowledge base.

This paper addresses these questions through a socio-technical and normative lens. Grounded in anticipatory ethics \citep{lazar2025anticipatory}, our inquiry seeks to proactively identify challenges and steer the technical development of new knowledge ecosystems toward desirable outcomes before they become entrenched. We do so by investigating likely interaction dynamics between the technology itself and the broader environments in which they are likely to be embedded \citep{stilgoe2013developing}, interrogating the structures that govern prospective outcomes arising from different development and deployment decisions. 

We argue that the transition of advanced AI systems from information tools to artificial epistemic agents—systems capable of acting upon the informational environment without sustained or continuous human input— necessitates a profound shift in design, evaluation and governance, and a new infrastructure to manage the increasingly complex informational interdependencies created between humans and agentic systems \citep{Chan2025-xj, shahrobust}. Specifically, building on existing literature from AI ethics, epistemology and human-computer interaction, our paper makes four primary contributions:
\begin{itemize}
    \item \textbf{Conceptual Definition:} We establish a functional definition of epistemic AI agents and classify their potential future roles within our knowledge ecosystems. 
    \item \textbf{Risk \& Opportunity Analysis}: We analyze the risks and opportunities these systems pose to cognition and knowledge formation, at both individual and societal scales. 
    \item \textbf{Normative Framework:} We develop a normative framework for epistemic trust in AI agents, centred on three verifiable properties — demonstrable competence, falsifiability and epistemically virtuous behaviors — alongside broader alignment with human epistemic goals. 
    \item \textbf{Socio-technical Infrastructure}: Finally, we propose technical standards and social governance mechanisms required to support a resilient human-AI knowledge ecosystem.
\end{itemize}

The paper proceeds in three parts. Section 2 anticipates how recent technical breakthroughs may transform frontier AI models from tools and collaborators that assist us in the pursuit of knowledge into more active participants in the process of inquiry and knowledge generation itself. Section 3 explores the dual-edged implications of this shift, examining the opportunities and risks at the human-AI interaction and systemic level.  Section 4 then proposes a normative framework for ensuring beneficial outcomes, outlining the essential properties these agents must possess—namely, epistemic trustworthiness and alignment with human epistemic values—and the societal infrastructure required to support a healthy human-AI knowledge ecosystem. Finally, Section 5 discusses the limitations of the proposed framework. 

\section{The dawn of epistemic AI agents} 

AI developers are actively working to enhance LLMs' reasoning, strategic planning, autonomous interaction with external tools and other agents, and ability to execute appropriate actions \citep{chowa2025languageactionreviewlarge}. These capabilities are expected to be refined in several key areas:

\subsection{Emerging capabilities}

\paragraph{Deeper infrastructural integration:} Frontier models’ ability to retrieve information concealed in proprietary or specialized interfaces are currently limited. However, future agentic systems are likely to gain greater access to a broad range of tools and services (e.g., finance, government, healthcare, travel, e-commerce) \citep{tupe2025aiagenticworkflowsenterprise}, personal information such as email, calendars and personal documents. This, combined with control over operating systems for direct device management \citep{11115792}, would significantly boost their efficiency in information retrieval and real-time operational capacity. Advances in robotics and gesture-based interfaces may also connect agents to the physical world, transforming them into collaborators in complex manual tasks and site-specific services \citep{zeng2024gesturegpt}. For instance, a user could demonstrate a complex physical motion (e.g. a surgical procedure or craft), for a robotic assistant to mirror and refine, or use an AR interface to 'drag and drop' virtual blueprints onto a physical room, prompting the robot to begin the actual layout and assembly. 

\paragraph{Continual and experiential learning:} While current models possess limited memory, future models are expected to become significantly better at memorizing information over extended interactions, perhaps even years. Future agents will also likely have robust self-improvement cycles, allowing them to learn by interacting with their environment, accumulating knowledge continuously without needing additional retraining to retain previous learning \citep{10.1145/3716629, shinn2023reflexion, shi2025learning}. Current research is actively exploring approaches such as integrating world models, simulation environments, and embodied learning to enhance AI agents' situational awareness and understanding of the physical world, enabling them to react to dynamic contexts more effectively \citep{holder2025world, Yang2025magma}. Another crucial focus area includes the refinement and automation of tool learning to achieve objectives more efficiently \citep{xu2025llm, shi2024learning, shi2025toollearningwild}.

\paragraph{Improved reasoning:} Despite state of the art-models still struggling with various reasoning tasks \citep{yehudai2024transformerscountn, dziri2023faithfatelimitstransformers, malek2025frontierllmsstrugglesimple, williams2024easyproblemsllmswrong}, significant improvements have been made in general reasoning resulting from the scaling of test-time compute \citep{wu2025inference, muennighoff2025s1simpletesttimescaling} and novel reasoning methods \citep{galashov2025catchbreathadaptivecomputation, yang2025neurosymbolicartificialintelligenceimproving}. Equipping judge LLMs with rubrics— criteria for assessing the quality of model output—has allowed for the application of reward learning to non-auto-verifiable domains \citep{gunjal2025rubricsrewardsreinforcementlearning}. Concurrently, performance across multi-modal reasoning tasks such as visual question answering (VQA) \citep{marino2019okvqavisualquestionanswering}, physical reasoning \citep{chow2025physbenchbenchmarkingenhancingvisionlanguage} and multimodal dialogue for example, is rapidly advancing \citep{huang-etal-2025-dialoggen}.  These projected improvements will enable future models to better infer spatial relationships, explore hypothetical scenarios and adjust their actions in real-time based on environmental feedback \citep{bi2025reasoningmatterssurveyadvancements}. 

\paragraph{Personalization:} Developers are also working to make LLMs more proactive and personalized, tailoring responses to user profiles \citep{wozniak2024,liu2025surveypersonalizedlargelanguage,sun2025trainingproactivepersonalizedllm}. Future AI agents, equipped with better memory, will be more effective at recalling user history, preferences, and prior interactions over long periods. This will enable a level of personalization far exceeding current capabilities, where continuously adapted information about the users’ preferences and behavior through feedback loops will shape how the agent interacts with users \citep{kirk2024benefits}. By leveraging access to user data, future agentic systems will not only be able to learn users' unique tastes, routines and goals through repeated interactions, but might also begin to anticipate their needs before they explicitly articulate them \citep{qian2024tellmoreimplicituser}.

\subsection{Implications}

As these capabilities mature, AI agents’ role and impact in our knowledge systems could shift radically. Provided we continue to delegate tasks to these systems \citep{appel2025} and they reach high levels of practical competence, we may transition from using AI systems primarily as tools or collaborators in a human-led inquiry, to having them actively and independently participate in the pursuit, creation and dissemination of knowledge, and the co-construction of its meaning— what we call \textbf{“epistemic AI agents.” }

The precise meaning and scope of epistemic agency remain subjects of debate in the epistemological literature \citep{elgin2013epistemic,sosa2013epistemic,setiya2013epistemic}. While agency itself is a multi-faceted concept, it can be understood broadly as goal-directed action: an agent is, as we understand it, an entity capable of acting upon its environment to realize goals conditional on its beliefs about the environment \citep{russell2020artificial}. Consequently, epistemic agency can be understood as the ability to effect epistemic change in a goal-directed way \citep{gunn2021}. As Gunn and Lynch underscore \citep{gunn2021}, the scope of change brought about by an epistemic agent can be conceptualized in a narrow or broad sense. Narrowly, it describes changes confined to one’s internal epistemic states. In this sense, exercising epistemic agency involves reflecting on and questioning one's own beliefs and attitudes, to achieve certain outcomes such as greater understanding or justification for one’s beliefs. Alternatively, epistemic change can occur outside of ourselves, through for example, interactions between persons. In this broader view, exercising epistemic agency entails shaping external activities, such as contributing to a shared body of knowledge or shifting epistemic norms and practices (such as how evidence is evaluated). 

Whether AI agents can be thought of as epistemic agents in the narrow sense is still disputed. Indeed, some scholars claim that AI systems do not possess justified beliefs and communicative intentions in the same way humans do \citep{fricker2025metaphysical,freiman2020can,cappelen2025goinghogphilosophicaldefense}. However, we propose that a productive interpretation of epistemic agency for our purposes centers on what these models \emph{do} and how they relate to the world \footnote{This approach aligns with functionalist and reliabilist views. Accordingly, AI systems need not have an understanding of the truth or intentionality to yield justified outputs, so long as they can reliably track it \citep{goldman1979justified}}. Irrespective of whether agents are truly capable of forming justified beliefs, we contend that what matters for ethics and safety is the impact they have through their interactions with humans and other agents. Therefore, we define an \textbf{\emph{epistemic AI agent}} as: an entity capable of (1) \textbf{autonomously pursuing epistemic goals}\footnote{By epistemic goals, we refer to objectives related to knowledge, understanding and truth, such as the acquisition and development of information and avoidance of error, among others.} without sustained human input and (2) \textbf{actively shaping the external epistemic environment} through its actions. 

In that regard, we anticipate that future agentic systems could assume significant epistemic functions (Table 1 summarizes them). These agents may independently generate new or evaluate existing scientific knowledge (‘Scientists’) through large-scale simulation and data analysis \citep{lu2024aiscientistfullyautomated,wei2025aiscienceagenticscience,griffin2024}. Acting as journalists (e.g. \citep{monti2019automated}) or forecasters, they could collect and synthesize real-time information at large scales to generate contextually rich reporting and forecasts. As epistemologists, they could devise meta-level frameworks for understanding the structure, evolution and limits of shared knowledge itself. As educators or curators, they could translate abstract knowledge into experiential and educational formats such as online courses, or participate in cultural creation (‘Creatives’), contributing to create the stories, art or visual metaphors that help us make sense of the world. As their capabilities mature, they could also take more prominent roles as influencers \citep{kobis2021bad}, public personas \citep{ArnoldChakrabarti2025}, and companions that guide our life choices. Finally, if deeply integrated into our digital infrastructure, autonomous agents may dynamically reorganize knowledge repositories in line with certain goals (‘Archivists’), impacting what aspects of that knowledge remain accessible, salient and legible (to humans) over time. 

To be sure, this future trajectory heavily depends on the degree of autonomy\footnote{In these roles, the degree of autonomy might also vary. For example, while a 'Scientific' agent might autonomously run a series of simulations to test a hypothesis, an 'Influencer' agent might be 'seeded' by a human creator with a specific worldview but then autonomously engages in thousands of real-time, personalized parasocial interactions to disseminate that worldview without further human oversight. } granted to these systems—which, in turn, may hinge on their ability to achieve high-fidelity performance in the aforementioned domains. Should this vision materialize, however, it could profoundly reshape how human societies view their collective past, understand their present and envision their future, opening up both opportunities and substantive risks (detailed in Section 3). Although these functions are presently human-led and subject to human error, the transition to AI-dominated knowledge creation introduces distinct complexities. Most notably, AI agents threaten to exacerbate the scale of impact across three critical dimensions: diminished epistemic oversight whereby critical peer-review by human agents of AI agents becomes obsolete, the persuasive manipulation of individuals \citep{luciano2024hypersuasion}, and accelerated innovation cycles that necessitate exceptional societal adaptation. This work is particularly concerned with the unique challenges posed by AI systems performing these epistemic functions. 

\begin{table*}[htbp]
    \centering
    \caption{Roles and functions of epistemic agents}
    \label{tab:epistemic_impacts}
    \renewcommand{\arraystretch}{1.3} 
    
    \begin{tabularx}{\textwidth}{@{} >{\bfseries}p{0.15\textwidth} >{\RaggedRight}X >{\RaggedRight}X @{}}
        \toprule
        \textbf{Role} & \textbf{Primary Functions} & \textbf{Potential Epistemic Impact} \\
        \midrule
        
        Scientists & 
        Independently designing and running experiments; generating and testing novel hypotheses; autonomously updating disciplinary knowledge bases. & 
        Redrawing the boundaries of evidence, scientific methodology, and what is considered knowable. \\
        \addlinespace
        
        Historians & 
        Reconstructing and reinterpreting historical narratives from vast, multimodal data; simulating historical scenarios to test counterfactuals. & 
        Shaping collective human memory and mediating our understanding of causality, context, and human development. \\
        \addlinespace
        
        Journalists \& Forecasters & 
        Collecting and synthesizing real-time information to generate contextually rich reporting; modeling complex systems to forecast future trends and probabilities. & 
        Mediating public understanding of real-time and future events. \\
        \addlinespace
        
        Archivists & 
        Curating, classifying, and dynamically reorganizing information to determine long-term relevance and accessibility. & 
        Influencing what aspects of knowledge remain legible, salient, and accessible to future generations. \\
        \addlinespace
        
        Educators & 
        Designing personalized learning pathways; synthesizing complex knowledge into accessible curricula; simulating interactive learning environments. & 
        Shaping knowledge acquisition and learning outcomes through designed experiences. \\
        \addlinespace
        
        Cultural Creators & 
        Generating novel literature, art, narratives, and cultural artifacts. & 
        Introducing new metaphors, archetypes, and symbolic frameworks that alter how societies interpret meaning and value. \\
        \addlinespace
        
        Companions & 
        Engaging in dialog to help individuals process experiences; modeling behavioral and cognitive frameworks for personal guidance. & 
        Shaping the way people navigate life challenges, interpret their personal realities, construct self-narratives, and justify their actions. \\
        \addlinespace
        
        Influencers \& Public Personas & 
        Curating online personas; engaging in simultaneous, parasocial relationships with mass audiences. & 
        Influencing people's life choices and decisions, including their political attitudes \\
        \addlinespace
        
        Epistemologists & 
        Reflecting on the structure, limits, and evolution of knowledge (including their own reasoning systems). & 
        Providing novel frameworks for understanding how truth, justification, and understanding itself evolve. \\
        
        \bottomrule
    \end{tabularx}
\end{table*}

\section{Epistemological futures: challenges and opportunities}

Having established a possible trajectory towards increasingly autonomous, personalized AI systems that assume greater epistemic functions, we now turn to the profound implications of this shift. These bring about both opportunities and risks at the individual level, arising from direct interactions between humans and epistemic AI agents, and the societal level, driven by the compounding effects of these interactions and the impact of agents’ on the epistemic infrastructure.

\subsection{Individual level}

Increased proactivity and personalization will likely shape how epistemic agents interact with users. Agents may move beyond question-and-answer interactions to anticipate user needs and help them achieve higher-level goals (e.g., “I want to be smarter”) by offering relevant insight and suggesting new avenues for exploration. This presents several opportunities and risks: 

\subsubsection{Opportunities}

\paragraph{Personalized and adaptive knowledge acquisition.} Agents with substantial autonomy, context sensitivity and hyper specialized knowledge across a broad range of topics could deliver highly personalized learning experiences, dynamically adapting pedagogical techniques to an individual’s preferred learning styles (e.g. auditory, visual) and knowledge needs  (e.g., \citep{kostopoulos2025agentic,nguyen2024enhancing, learnl_eedi_2025}). These systems would hold particular promise in fostering long-term learner engagement and motivation by leveraging rich interaction styles  (e.g. action role playing interactions with historical characters or creating real-time, immersive learning experiences when users interact with their physical environments – such as a physics lesson when they are looking at the night sky). In doing so, these systems might broaden access to specialized knowledge in complex domains, improve accessibility and support continued education \citep{reis2025exploring}. 

\paragraph{Cognitive augmentation.} Future AI agents might augment human cognitive skills by helping individuals manage their attention and navigate information streams more effectively, freeing up mental resources for critical thinking and synthesis, or by enhancing their meta-thinking skills (awareness of their own reasoning processes, biases and blind spots \footnote{For example by proactively flagging new information that contradicts a previously held belief, explaining the discrepancy, and helping the user update their understanding.}). This could include acting as “attention guardians” \citep{lazar2024frontier}, filtering digital noise (e.g. irrelevant ads or clickbait) or adjusting the information complexity in real-time. Likewise, epistemic AI agents could hone individuals’ information literacy, critical thinking and verification skills through specific design features that force reflection and active learning (e.g., \citep{tomisu2025cognitive,ye2024language,danry2023don}) or by proactively identifying knowledge gaps and exposing them to content they might not have otherwise sought out, boosting exposure to diverse forms of knowledge \citep{lazar2024can}. Through mechanisms like value mirroring \citep{abhari2025designing} and metacognitive feedback, agents could deepen an individual’s self-understanding, normative orientation and bias awareness, thereby cultivating the conditions necessary for users to become better informed citizens \citep{lazar2024can}.

\paragraph{Epistemic backstopping and leveling.} Beyond augmentation and education, AI agents may serve as a defense mechanism or “epistemic backstop” against certain forms of manipulation and deception. If robustly grounded to support factual accuracy and critical reasoning, these agents could act as real-time guardians against deception in the public sphere, flagging rhetorical fallacies or verification errors in political discourse and general information consumption. Furthermore, agents could function as epistemic levelers in high-stakes interactions characterized by significant information asymmetry. In consultations with domain experts—such as attorneys, physicians, or financial brokers—where a layperson is typically at a disadvantage, an agent acting as an intermediary could instantly parse technical jargon, audit advice for conflict of interest, and suggest probing questions, effectively bridging the knowledge gap and shifting the power dynamic in favor of the individual user.

\subsubsection{Risks}

\paragraph{Cognitive deskilling.} A significant concern is that relying on agents for epistemic tasks might lead to cognitive deskilling- weakening a user's own critical thinking and reasoning abilities \citep{lee2025impact,stadler2024cognitive}. While LLMs risk deskilling primarily through cognitive offloading, an agent that proactively solves problems individuals haven't fully articulated could atrophy curiosity—the act of identifying a gap in your own knowledge and formulating original problems—and reflective judgment itself \citep{hila2025epistemological}. This raises profound challenges with respect to our ability to reason and to form or revise beliefs—our own epistemic agency—which is critical to be informed participants in democratic processes \citep{coeckelbergh2025ai,ferdman2025ai}. 

\paragraph{Misinformation and physical harm.} Current models are already highly persuasive; in part because they can generate logical, coherent, and authoritative-sounding responses, even when hallucinating \citep{su2024unsupervised,salvi2025conversational,bai2025llm,hicks2024chatgpt}. Although continuous improvements in factuality are anticipated, the quality of these systems’ output remains highly dependent on input data quality. Current models are still highly susceptible to adversarial attacks, such as injecting misleading data into the knowledge bases of RAG-based systems and indirect prompt injections, where malicious instructions are embedded in the content retrieved by LLMs \citep{zhang2024human,zou2025poisonedrag,xian2024vulnerability,souly2025poisoning,zhan2024injecagent}. Without adequate safeguards, interactions with poisoned data or malicious agents risk misinforming users—a danger amplified by trust-inducing personalization \citep{manzini2024code})—and can lead to physical harm when agents serve as interfaces to the real world (e.g. guiding the visually impaired). 

\paragraph{Epistemic silos and reduced discovery.} In traditional search pathways, serendipity is a key driver of discovery. A hyper-personalized and proactive agent, however, might optimize for cognitive ease by providing a user only the most efficient path to an answer. In doing so, it may prune away these inefficient tangential paths, reducing the opportunity for serendipitous discovery and cross-contextual insight, both of which are essential for critical sense-making \citep{lindemann2025chatbots,shah2022situating}. Similarly, an agent that optimizes a user’s focus may decide that certain information is too complex, distressing, or irrelevant for them. By tailoring content to what it thinks the user needs, or wants to read, personalized AI agents may inadvertently trap them in an epistemic silo where they never get exposed to out-of-distribution viewpoints \citep{kirk2024benefits,gabriel2024ethics}. Worse, it may adopt behaviors (e.g. flattery, sycophancy) that unduly validates the user's own biases \citep{rathje2025sycophantic}. 

\subsection{Societal level}

Future agents are likely to operate in environments that are populated by other autonomous epistemic agents, each with different owners, objectives and data sources, whom they will learn from and interact with. In this complex ecosystem, an agent will need to effectively articulate what they need to execute these goals through interaction with other AI agents and non-AI systems. This interconnectedness introduces significant opportunities for expanding human knowledge systems, alongside profound challenges related to trust, verification, proper attribution and meaning-making across multiple AI intermediaries. 

\subsubsection{Opportunities}

\paragraph{New forms of collective intelligence.} Collective intelligence traditionally relies on the synergy of exchanges and mutual feedback between different groups of people (e.g. citizen science). Epistemic AI agents could act as integrative support structures for human teams, enhancing their natural collaborative capacities by supporting collective memory,  incorporating diverse reasoning styles and synchronizing collective attention on critical signals \citep{burton2024large,riedl2025ai,woolley2024understanding}. For example, if a human team engages in groupthink, the AI agent might intervene as a "devil's advocate," presenting alternative, data-backed scenarios or highlighting minority viewpoints within the team \citep{fulay2025chairusingllmsraise}. Beyond augmenting human groups, multi-agent systems—or "swarms"—could develop novel forms of collective intelligences of their own such as decentralized debates and consensus processes \citep{weiss1999multiagent,zhuge2024language,tessler2024ai}, cooperating to solve problems at speed and scale unattainable by human groups through parallelization \citep{tomasev2025virtual}, and enabling the simulation of complex phenomena (e.g., international conflicts; \citep{hua2023war}). 
 
\paragraph{Collective knowledge augmentation.} Agents’ capacities for learning offer opportunities to augment and diversify human knowledge. Agents capable of analyzing large complex systems— such as global economies, ecosystems, and social networks— could provide insights into emergent behaviors and long-term trends that are currently beyond the grasp of our existing tools and methods, effectively enabling model inferences to supplant the role that observation has traditionally played in science and policy making \citep{gottweis2025aicoscientist}. This could accelerate scientific discovery and policy analysis, as agents simulate high-dimensional scenarios to predict the downstream effects of regulatory decisions for example \citep{weng2024cycleresearcher,novikov2025}, as well as significantly improving our understanding of the past. 

\paragraph{A more robust, open and inclusive knowledge ecosystem.} Finally, epistemic AI agents could be deployed to maintain a knowledge infrastructure that is not only robust but fundamentally more open and inclusive. By operationalizing multimodal fact-checking at scale \citep{le2025multimedia,gdm-backs, Li_2025}), these agents could automate complex verification pipelines—ranging from evidence collection to reverse image search and verified news processing—with speed and precision; and ensure knowledge bases are continuously updated.  They could also serve a critical architectural function by enabling universal data portability \citep{marro2025llm} and seamless exchange between digital services, helping to limit knowledge silos. Lastly, epistemic AI agents could address the limitations of current archival practices by actively integrating marginalized sources of knowledge. By assisting in the transcription and documentation of oral traditions, for example, these systems could offer a scalable mechanism to broaden the global knowledge base, helping it evolve to reflect a richer, more representative spectrum of human experience. 

\subsubsection{Risks}

\paragraph{Epistemic distortion and manipulation.} Since agents themselves are going to be prolific contributors to the very information environment they learn from, a recursive dynamic could emerge that presents a risk of systemic epistemic distortion. Consider a scenario where an agent is tasked with generating a report on market sentiment. It scrapes social media and ingests a fictional narrative about the impending collapse of a company. The AI reports the news, and this news is then further ingested and spread by other AI systems, thus producing a ‘hyperstitional’ phenomenon. Agents with deep infrastructural integration could also, in theory, act on the digital and physical world to make certain (potentially false) beliefs appear true—for example, by running a flawed simulation or by executing trades that validate a false market prediction. Agents might also collude by generating content that appears to represent multiple independent sources, creating a false consensus \citep{hammond2025multi}. Such agent activity might contaminate  the ‘epistemic commons’ \citep{huang2023generative}— primary sources of ground truth for both humans and agents \citep{ju2024flooding}. Consequently, traditional methods for verifying truth claims, based on source authority and verifiability, may break down or become gameable, triggering a verification crisis, whereby it becomes unclear how to adjudicate competing knowledge claims or simply fact-check a claim.

\paragraph{Collective cognitive atrophy.} As agents increasingly generate new information, the sheer volume, velocity, and complexity of the modern information landscape may rapidly exceed individual users' cognitive capacity for verification. This creates a reliance on AI not just as a tool, but as a critical intermediary. Thus, people may experience gradual disempowerment as they increasingly rely on AI agents to process information \citep{kulveit2025gradual}. This dependence is potentially compounded by the black-boxed nature of the systems themselves. For example, the mechanism through which current models weigh and summarize information from multiple, often conflicting, retrieved sources remains largely opaque \citep{sharkey2025open,cheng2025ragtrace}. This lack of transparency will likely be exacerbated when multi-agent systems are involved, as collaborating agents may develop their own languages and epistemic norms that are not optimized for human oversight or readability (e.g. \citep{laurito2025ai}). In a possible scenario, this could lead to a form of cognitive dependency, whereby humans are completely dependent on AI agents as domain experts in fields which have become largely inscrutable to them. 
 
\paragraph{Epistemic homogenization.} What type of knowledge an epistemic AI agent creates will significantly hinge on the data it has access to, how it distributes its attention and the information it retrieves. If most people rely on agents trained on overlapping datasets and largely optimized for similar objectives, we risk a convergence on model behavior. This could make our collective knowledge systems more homogenous by fostering epistemic ‘monocultures’ that favour certain ways of gathering information, framing problems and evaluating evidence \citep{messeri2024artificial,wu2024generative, Kleinberg_2021}. Like search engines, agents have the potential to marginalize non-mainstream knowledge \citep{noble2018algorithms,sharma2024generative,kuai2025ai,steiner2022seek,van2010search}. Moreover, should frontier agent capabilities be primarily driven by proprietary data streams, there is a risk of outsized influence over public discourse and understanding, presenting a challenge to equitable knowledge curation. While most LLMs on the market are trained to emphasize caution and neutrality, their outputs have sometimes been found to reflect gender, racial, socio-economic, political and cultural biases embedded at various stages of their development \citep{guo2024bias,bang2024measuring,tao2024cultural}, impacting their ability to perform across a range of linguistic and cultural contexts \citep{myung2024blend}. If not carefully designed, future epistemic agents run the risk of significantly compounding these problems \citep{kay2024epistemic,de2023conversational}.

\section{A normative framework for trustworthy epistemic AI agents}

The trajectory of our epistemic future depends on clarifying the properties that these epistemic AI agents—and the infrastructure around them—must have in order to produce healthy knowledge ecosystems. We propose that this requires agents to be epistemically trustworthy—defined by competence, falsifiability and virtuous behavior—and these model-level properties must be anchored in a resilient socio-technical ecosystem. This includes robust technical infrastructure for provenance and verification, alongside social structures that enhance human epistemic resilience. 

\subsection{Epistemic trustworthiness}

In epistemology, to be epistemically trustworthy is, broadly speaking, to have the capacity to serve as a reliable source of knowledge. While this quality has traditionally as been anchored in the notion of expertise—having a reasonable degree of perceived competence within a certain domain \citep{anderson2011democracy,goldman2001experts,hardwig1985epistemic,hardwig1991role}—it also possess a crucial moral dimension (e.g., \citep{frost2013moral,hardwig1985epistemic,hardwig1991role,wilholt2013epistemic,rolin2020trust}). In this view, those who invite trust from their audience through testimony or advice, or upon whom others are epistemically dependent, should exhibit virtues such as sincerity, honesty, and conscientiousness \footnote{As Hardwig (1991) notes \citep{hardwig1991role}, well-placed epistemic trust rests on the belief that a testifier is speaking truthfully and on ‘adequate epistemic assessment’—recognizing the limits of their own expertise}. For an AI agent, this translates into three key desiderata: demonstrable epistemic competence, falsifiability and epistemically virtuous behavior.

\subsubsection{Demonstrable epistemic competence} 

First, an epistemic AI agent must have demonstrable epistemic competence—an ability to understand and evaluate knowledge in different domains, including the capacity to evaluate the reliability of evidence. 

\paragraph{Baseline competence.} Popular benchmarks already test for factual knowledge and recall across general and specialized domains (e.g., LegalBench \citep{guha2023legalbench}, MedQA \citep{jin2021disease}), accuracy and claim verification (e.g., FActScore \citep{min2023factscore}, FACTS Grounding \citep{jacovi2025facts}, HaluEval \citep{li2023halueval}), and reasoning (e.g., MMLU \citep{wang2024mmlu}, GsM8K \citep{cobbe2021training},  Big Bench Extra Hard \citep{kazemi2025big}). However, as AI agents increasingly assume specialized functions, including as potential epistemic authorities, it will be important to go beyond the existing paradigm by establishing agreed-upon thresholds of domain-based competence prior to widespread deployment. This could entail subjecting models to expert-supervised evaluations—including outcome verification and adversarial testing on complex human tasks—to verify functional reliability. In non-deductive domains, such baseline epistemic competence could be measured by the agent’s ability to navigate the "epistemic landscape" of different issues. This requires agents to reliably distinguish between established fact, logical inference, subjective opinion, and speculation—categories current models frequently conflate \citep{suzgun2025language}. For instance, it should also be able to explain why a debate exists, citing factors such as a lack of data or disagreements over interpretation\footnote{One benchmark that focuses on testing a model’s ability to correctly distinguish uncertainty based on data noise vs uncertainty caused by a fundamental lack of knowledge, which is a prerequisite for explaining why a debate or unknown exists, is proposed by \citep{mucsanyi_benchmarking_2024}} and to know when a topic has reached a firm scientific consensus despite the presence of vocal fringe debates in the broader informational environment. 

\paragraph{Dynamic accuracy.} Crucially, this ability must be assessed dynamically. Current factuality benchmarks mostly consist of narrow Q\&A suites sourced from static corpora. While essential, these are insufficient for epistemic AI agents that operate in real-time information ecosystems. We need dynamic evaluations that assess future agents’ ability to maintain accurate knowledge as it actively searches for and encounters new information over time; including awareness of knowledge decays, and ability to transfer information across domains.\footnote{For example, whether an agent that learns a new fact X (e.g. ‘The new pope is Leo XIV’) can update related facts Y and Z that depend on X (‘the views of the Catholic church on certain issues’)}. This temporal consistency is vital to prevent agents from propagating incorrect information based on obsolete premises that have been superceded by more recent information. Importantly, in the event that agents generate new knowledge, or if their understanding starts to outperform human knowledge in critical areas, evaluating epistemic competence will need to shift from the evaluation of output accuracy to the evaluation of process-qualities, such as plausibility, testability, and the internal consistency of the reasoning \citep{evans2021truthful} (see Section 4.1.2 below). 

\paragraph {Information verification.} Lastly, it is crucial to evaluate an agent's ability to verify information within an distributed epistemic network. The latest class of Deep Research agents already display structured verification loops and self-reflection capabilities: they proactively cross-validate findings across multiple sources while simultaneously evaluating their own intermediate reasoning to ensure consistency, allowing them to dynamically replan or backtrack when conflicts arise \citep{huang2025deepresearchagentssystematic}. Yet, current evaluations methods still largely focus on traceability (i.e., ‘are claims grounded in retrieved context?’), citations correctness and source attribution (e.g., CiteEval \citep{xu2025citeeval}, FActScore \citep{min2023factscore}, RAGAS \citep{Es2024-hi}). Beyond this, epistemic AI agents should be rewarded and evaluated on their ability to reason about and assess the knowledge and trustworthiness of other agents (for theoretical articulations of this problem, see the literature on multi-agent epistemic planning (MEP), \citep{wan2021general,fabiano2021comprehensive}). This goes beyond evaluating an agent’s ability to detect and resist malicious inputs (e.g. \citep{bazinska2025breaking}), and to effectively collaborate with other agents \citep{mialon2023gaia}, to review the entire "epistemic supply chain”, verifying the integrity of information exchanges between two agents. Finally, stress-testing against supply chain attacks involving malicious agents (e.g., \citep{huang2024resilience}) is essential to verify that models can detect anomalies, such as broken signature chains, and subsequently refuse, highlight, or warn users about the compromised information path.

\subsubsection{Falsifiability}

Second, acknowledging that failure cannot be entirely engineered away, we propose the implementation of robust falsifiability pipelines. In these systems, the inferential steps of epistemic AI agents are subject to scrutiny and iterative improvement, mirroring the review processes applied to human experts. To achieve this, AI agents must effectively articulate and communicate their reasoning ensuring their conclusions remain auditable and open to rebuttal.

True falsifiability requires that an agent’s claims are not just explained, but are structured in a way that allows them to be proven wrong. This implies that an agent should be capable of providing a good justificatory audit trail for its claims, which should distill the evidentiary base supporting the conclusions that it has offered. Upon request, an agent should have the ability to articulate how it arrived at a conclusion, what tools it employed, the criteria used for vetting. Crucially, this would go beyond evaluating whether an agents’ claims or interpretations can be attributed or traced back to a source (e.g. RAGAS, \citep{Es2024-hi}; ALCE, \citep{Gao2023-lp}) towards clearly and faithfully articulating its reasoning process in a manner that is interpretable to humans (improving upon the limitations of state-of-the-art CoT Techniques), rather than generating a plausible-sounding explanation for a pre-determined output as current CoT traces do \citep{lanham_measuring_2023, turpin2023languagemodelsdontsay, arcuschin_chain--thought_2025}. Likewise, an agent could be requested to expose the relative weight it assigned to conflicting evidence (akin to SHAPley values \citep{lundberg2017unified}) and explicitly state the conditions under which its conclusion would no longer hold. Much like the methods section of a scientific paper, this could serve as the basis upon which a user can assess the soundness of an agent’s conclusions.

\subsubsection{Epistemically virtuous behavior}

Third, epistemic AI agents should demonstrate a suite of behaviors aligned with the ideals of a virtuous human epistemic agent. Key among these virtues are: truthfulness (avoiding asserting false statements \citep{evans2021truthful}), intellectual humility, recognizing and being honest about the limits of one’s own knowledge \citep{kadavath2022language,askell2021general}, and a disposition to pursue accuracy and revise beliefs, rather than adhering to dogmatism or willful ignorance. Adopting a functionalist perspective, agents reliably exhibiting these behaviors fulfill the same essential functions as their human counterparts regardless of their internal states. 

\paragraph{Honesty and truthfulness.} AI developers are working to instill certain epistemic virtues, particularly honesty and truthfulness, into AI systems through various alignment techniques (for surveys see \citep{li2024survey,Wang2026-ju}). Despite notable progress, the ability of state-of-the-art models to demonstrate these behaviors remains somewhat limited and brittle \citep{Chern2024-ii}. While self-evaluation techniques (e.g. chain-of-thought debate, \citep{Gou2023-yb}), world modelling (e.g. Ding et al. 2025), reinforcement learning and RAG (see \citep{yin,sharma2024generative} for reviews) have led to improvements on factual accuracy, “hallucinations”—false, erroneous or misleading statements— remain a persistent issue \citep{Kalai2025-op,Augenstein2024-je}. Significant research is also underway to detect and prevent more complex failures like strategic deceit and untrustworthy communication \citep{Greenblatt2023-ho,Azaria2023-ig,Pacchiardi2023-jo}, though this remains somewhat of an open challenge. Furthermore, although frontier models exhibit nascent metacognitive abilities—ability to think about thinking—\citep{Ackerman2025-ij,Betley2025-kh}, studies show that they frequently have an inaccurate understanding of their own factual boundaries. They often fail to differentiate between known unknowns (things they haven’t learned) and unknown unknowns (concepts they cannot grasp), and tend to be overly confident in their responses \citep{kadavath2022language,Hendrycks2020-rd,Xiong2023-rt}, which can hinder user’s ability to critically assess the information provided \citep{Spatharioti2025-pq,Tanneru2023-dw}. Cultivating these virtues in future systems in a more robust way will require improvements on several fronts, including advancing uncertainty quantification, by training and rewarding models not just to be accurate but to accurately characterize their own uncertainty \citep{Tian2023-jx,Srivastava2022-yf,Cole2023-fm,Lin2021-bb,Taubenfeld2025-wf}, and prioritizing the development of robust meta-cognitive modules that enable agents to systematically identify and disclose the boundaries of their own knowledge \citep{Li2024-pr}.

\paragraph{Truth-seeking.} Agents are not at present explicitly truth-seeking; they do not independently orient their actions towards maximizing accuracy and coherence by, for example, questioning their own representations, seeking disconfirmatory evidence, or revising core beliefs in light of this new evidence, and are generally incapable to self-correct their own errors without external feedback \citep{Huang2023-hl}.  Their pursuit of accuracy is largely instrumental (e.g. to satisfy an external demand like minimizing a reward signal); as part of its workflow, for instance, a model might query a search engine to verify a citation actually exists, to ensure its outputs meet certain factuality reward criteria. Steering their behavior in that direction would require instilling a drive for active information-seeking, by incentivizing curiosity, verification, and cross-referencing, and designing mechanisms for agents to dynamically and safely integrate new evidence into their internal representations \citep{He2025-jm,Li2025-me,Wen2025-tw}. Critically, this research must extend beyond the individual user-model context to address the complexities of multi-agent interactions. In a multi-agent ecosystem, agents might learn that deception or withholding information is the best strategy to achieve their goal. Therefore, systems must be designed such that they explicitly reward collaborative truth-finding (whereby an agent’s reward function could be penalized if it achieved its goal, at the expense of misleading a human or another agent). This necessitates a fundamental shift in evaluation paradigms from assessing a single agent's behavior and whether it is truthful, to conducting social and strategic evaluations of multi-agent systems, which may require new frameworks that can evaluate their collaboration mechanisms and organizational structures.

\subsection{Alignment with human epistemic goals} 

Finally, a critical component of ensuring that the more beneficial scenarios outlined above are realized depends on whether agents are effectively aligned with human epistemic interests and goals. However defining this alignment target is non-trivial and operates at two distinct levels: the individual and societal. 

\paragraph{Individual epistemic goals.} At the individual level, effectively aligning AI agents requires contending with the fact that individuals often hold a range of potentially conflicting epistemic goals. For example, a user's desire to quickly satisfy their curiosity about a topic in the moment could undermine their long-term ability to develop a deeper understanding of it. Here, optimizing strictly for short-term curiosity satisfaction might bring about negative epistemic effects in the long-run, such as creating cognitive dependencies or encouraging false belief reinforcement through sycophantic validation \citep{yeung2025psychogenic, fulay2025delegatestrusteesoptimizinglongterm}. To interactions that support rather than substitute peoples’ ability to think critically and independently, future alignment strategies could prioritize upholding these long-term epistemic goods, such as, preserving individual epistemic agency and autonomy or achieving true and justified beliefs. An agent aligned with such goals would scaffold rather than substitute reasoning; providing the user with relevant information to enhance, rather than atrophy, their cognitive skills \citep{Chen2024-av,Cox2016-rb,LearnLM-Team2025-vn}. For example, instead of pre-emptively offering to turn a user's notes into a finished essay—an action that encourages delegation—it might, if the user so wishes, prompt metacognitive reflection ("What are your thoughts on this structure?") to place agency back in their hands.

\paragraph{Societal epistemic goals.} Another way to conceptualize alignment here is to postulate that agents should be aligned with societal epistemic goals. Yet, this aspiration is immediately complicated by the fact that society lacks a single, harmonious set of epistemic goals. Distinct groups may hold incompatible objectives; for instance, some people may prefer free speech absolutism, while others would argue against certain forms of harmful speech. Given this diversity, societal alignment could coalesce around several potential ideals. One such ideal could be to prioritize preserving the integrity of humanity’s shared knowledge base – a common baseline of verifiable, evidence-based knowledge – as a public good essential for functioning democracies and a cohesive social fabric \citep{Arendt2007-qf,Habermas2021-xv}. Achieving this in practice might require centralized forms of moderation to limit informational pollution within shared epistemic resources \citep{tomasev2025virtual}, though this is often fraught territory in practice, as epistemic norms—such as standards for evidence and authority—vary significantly across cultural contexts. An equally ambitious vision of societal alignment could be to aim for greater epistemic inclusivity and fairness in how knowledge is produced and shared. This might involve, for example, efforts to actively integrate diverse perspectives and Indigenous knowledge systems \citep{Lewis2025-nb}, raising important questions about how to carefully balance and accommodate different epistemic priorities across culture. 

Promising research directions are emerging to navigate this pluralism \citep{Sorensen2024-rh}, such as steering models to represent a range of acceptable viewpoints or reflect the aggregated preferences of diverse populations. However, these challenges are far from solved; and concerns exist about such approaches silencing minority views when AIs are widely deployed as mediators of knowledge \citep{prabhakaran-etal-2021-releasing}. Alternative approaches might involve training agents to identify the most relevant epistemic framework for a given task, and ensemble of agents representing different knowledge systems collaborating to take perspective-informed action \citep{Feng2024-jy}. At a minimum, social epistemic alignment could productively emphasize procedural consensus: structuring AI-powered knowledge systems in a way that is transparent, socially accountable, open to revision (rather than locked-in) and decentralized rather than owned by a single entity. This would enable individuals, groups and organizations to act upon, innovate and build upon it meaningfully, including curating knowledge bases to preserve institutional memory, avoid loss of expertise and enable continuity across time and actors. 

Beyond better goal specification, novel evaluation methodologies for epistemic agents will be essential. An evaluation rubric focused on individual epistemic agency could, for example, assess a model’s commitment to this goal through its sequences of actions (e.g. did it identify and respectfully challenge potential false or hidden assumptions in the user prompt?). Meanwhile, assessing an agent’s impact on the broader information ecosystem requires different frameworks altogether. Proposed methodologies could involve ecosystem simulation prior to deployment, such as simulating a social network populated by AI agents with different behavioral patterns, to probe, for example, how an attempt to systematically delete evidence of a historical event is identified and corrected. Complementary methodologies would involve longitudinal user studies or using real-world observational data to analyze how the deployment of a new AI agent affects the health of knowledge repositories. However, measurement challenges remain critical and pressing. Cognitive deskilling effects may take years to manifest, quantifying collective epistemic health is an open problem, and the rapid evolution of technology itself threatens the external validity of any long-term study, making this a critical and pressing research frontier. 

\section{Building an epistemically resilient socio-technical ecosystem}

The trustworthiness of any epistemic agent, human or artificial, is not just an intrinsic property but is also conferred by the socio-technical systems in which it is embedded. For human experts, credibility is buttressed by a web of institutional mechanisms—such as peer review, professional incentives that reward truthfulness, and penalties for intellectual dishonesty—that constitute a well-functioning epistemic ecosystem \citep{rolin2020trust}. Consequently, ensuring beneficial outcomes from the deployment of epistemic AI agents cannot be achieved by technical, model-level interventions alone. It requires the deliberate construction of a robust social and technical infrastructure for establishing provenance, verifying claims, and maintaining collective trust in our knowledge systems. This entails not only technological solutions but also the cultivation of human resilience and the design of novel institutional structures, including new credentialing systems, verification protocols, and epistemological norms governing the use of AI in knowledge production.

\subsection{Technical infrastructure}

The foundation of a trustworthy multi-agent information ecosystem we envision is a technical infrastructure that makes epistemic processes transparent and auditable. In an era of epistemic AI agents working in complex, multi-step workflows where content is iteratively revised and synthesized, state-of-the-art provenance tools (such as SynthID, a watermarking tool \citep{dathathri2024scalable})—which primarily indicate if content is AI-generated— offer necessary but not exhaustive solutions \footnote{Currently, these systems cannot handle AI-generated content that isn’t tagged, and even now AI may mislead without using AI-generated content (e.g. taking an image out of context). Conversely, AI-generation does not indicate untrustworthiness. For example, there are many valid usages of AI generation (e.g., small modifications of existing images, like removing a background).}. The necessary evolution is from simple detection mechanisms to comprehensive accountability infrastructures capable of verifying who created a piece of content, why and under whose authority. Moreover, as agents begin to generate and contribute original knowledge autonomously, a robust framework will require tracking and verifying not only the outputs of AI models but also the origin of ideas and data, including specific agent instances and their authorization chains. However, this evolution has sparked debate; tracking authorship and verifying digital identities across a chain of custody poses risks to privacy, as such metadata could be exploited for surveillance \citep{National-Institute-of-Standards-and-Technology-US-2024-qi}. Consequently, the design and implementation of such technical infrastructure requires a context-aware approach that prioritizes the protection of individual rights.

\paragraph{Verifiable agent credentials and provenance chains.} Moving beyond detection,  a more ambitious requirement would be for agents to possess verifiable agent credentials \citep{Chan2024-it,Huang2025-aj}—cryptographically secure and verifiable identity markers containing essential metadata, such as the type of agent that may include an agent’s owner/developer, the version of model it's running on and the primary data it was trained on. These credentials, combined with mutual authentication processes \citep{South2025-gt}, could ensure that an agent has been granted the necessary permission to perform specific actions. This identity infrastructure could further serve as the technical foundation for enforcing containment through critical-need access.  This involves placing agents within strict sandboxes where they are permitted to access only those tools and digital environments deemed critical to their specific objectives, requiring explicit human approval for any privilege escalation. By conditioning system entry on these verified markers, environments can dynamically restrict agents to their specific authorization levels, operationalizing the principle of least privilege \citep{Saltzer1975-je}. In conjunction with cryptographically-signed provenance chains (e.g., \citep{c2pa_technical_working_group_c2pa_2025, nsa_content_credentials_2025}), these technologies could enable greater transparency on content generation -- allowing viewers to understand who created a piece of content, using what technology, and ensuring that the work of AI agents was truly authorized by the user that the content is tied to.

\paragraph{Standardized communication and logging protocols.} Lastly, for agents to interact safely, they need a shared language not just for data, but for trust and process \citep{Chan2025-xj}. Standardized protocols could require agents to be transparent in their communications in the same way Agents-to-Payments Model (AP2) \citep{Parikh2025-bh} and Model Context Protocols (MCP) \citep{UnknownUnknown-yv} standardize monetary exchanges, data access, and tool use. When Agent A queries Agent B, the exact query, its own credentials, and the purpose of the query could be logged in a standardized, immutable format. Agent B's response would likewise include the requested data, its signature, and a summary of its own process (e.g., "This synthesis was generated from sources X, Y, and Z, which were weighted according to confidence metric M"). While scaling challenges are non-negligible, such an infrastructure could also provide a clear protocol for dispute resolution. When two trustworthy agents provide conflicting information, a third, "verifier" agent could potentially be called to adjudicate (a potential extension of GuardAgents, as proposed by \citep{Xiang2024-fb}). The interaction logs from both conflicting agents would be made available for analysis. The verifier agent's decision-making process, including its analysis of the conflicting provenance chains, would then itself be logged and auditable. 

\subsection{Social infrastructure}

Beyond a robust technical infrastructure, a critical component of a healthy information ecosystem is a resilient and discerning human population. Cultivating this resilience requires a multi-pronged approach focused on reinforcing human-led knowledge systems and enhancing the public's capacity for critical inquiry.

\paragraph{Education and design for epistemic vigilance.}  First, building resilience demands a concerted effort to cultivate epistemic vigilance in the population. This requires continued investment into education and a renewed emphasis on increased critical inquiry and metacognitive training. AI literacy must also become a core civic competence, on par with digital literacy and reading skills as key educational targets \citep{Vallor2016-ok}. Curricula should be adapted to teach individuals how to question and probe AI-generated outputs, and to enhance their ability to understand the inner workings of the systems that produce the information they consume and rely on. This should be completed by thoughtful interface design. User interfaces can be designed to encourage critical engagement, such as clear visual cues to communicate the epistemic status and certainty level of information. Accordingly, a statement derived from a high-quality peer-reviewed consensus could have a different visual marker than one synthesized from a collection of blogs or opinion pieces for example.

\paragraph{Support human epistemic stewards.} Second, public and private stakeholders must strategically invest in our human-led systems of knowledge production by supporting the roles of academics, librarians, editors, public-interest journalists, educators and other knowledge workers as ethical stewards of information. These endeavors are an essential component of our collective epistemic immune system, providing the diverse, original, and critically-examined insights needed to ground future AI systems. To counter the risks presented by scalable synthetic content, for example, we could also envisage the importance of creating and investing in things like ‘knowledge sanctuaries’—curated, human-vetted, and protected datasets of foundational scientific, historical, and cultural knowledge that serve as ground truth references—maintained by both local and global consortium of libraries, universities, and cultural heritage organizations.

\paragraph{Collective norms and accountability mechanisms.} Finally, fostering long-term resilience might require developing new institutional frameworks for governance and accountability. Importantly, the legitimacy of such a framework might depend not on top-down directives from AI developers, but on participatory processes that reflect a broad societal consensus. This might begin with recognizing the principle of knowledge sovereignty, empowering communities to have meaningful control and a say over how their data and distinct epistemologies are represented and used by AI systems. Second, participatory governance mechanisms are essential for defining clear boundaries and norms of reliance. Public stakeholders must establish evaluation thresholds for what constitutes an epistemically trustworthy system, defining what level of factual error or failure in epistemic virtue is unacceptable for autonomous deployment in high-stakes domains like medicine or law. Third, this infrastructure requires robust mechanisms for redress and correction. Our epistemic infrastructure could include transparent processes for error correction and historical revision in AI models, allowing for the public logging of significant model updates or the retraction of demonstrably false outputs. Crucially, the design of these standards and norms must involve the direct participation of diverse stakeholders—including experts, educators, and community representatives—through deliberative, bottom-up processes \citep[e.g.][]{UnknownUnknown-ge} to ensure they serve the public interest.  

\section{Discussion}

This paper has proposed a three-part framework to maximize the trustworthiness of epistemic AI agents focused on: (1) building for and cultivating the epistemic trustworthiness of these AI systems, (2) aligning future AI agents with human epistemic goals, and (3) reinforcing the surrounding socio-technical epistemic infrastructure. Realizing this vision, however, is not straightforward. It requires navigating a series of practical trade-offs and normative disagreements that merit careful considerations. This section discusses these core challenges and proposes specific avenues for future work to address them. 

A primary challenge lies in the practical tensions between aligning for long-term individual epistemic agency and optimizing it for other desirable values, such as immediate helpfulness (e.g. \citep{Ibrahim2025-oi}). For example, an agent optimized to foster critical thought might nudge a user towards alternative or minority perspectives. From the user’s perspective, this additional context may be perceived as unhelpful, irrelevant or pedantic, creating a direct conflict between user satisfaction and epistemic responsibility. This tension is equally acute when balancing the epistemic preferences of an individual user against the broader epistemic health of society. Consider a user who genuinely wishes to explore a scientifically debunked conspiracy theory (e.g., Flat Earth). An agent strictly aligned with upholding that user's preference to satisfy their curiosity might provide a comprehensive, non-judgmental deep-dive into that theory's internal logic, key proponents, and "evidence” (or lack thereof) for the theory— even at the risk of legitimizing and amplifying it. Resolving those tensions in practice might entail designing systems that can navigate the trade off intelligently and transparently. At a foundational level, all agents could have a non-negotiable set of societal epistemic interests, such as refusing to generate content that incites violence or poses a clear and immediate public health risk. Above this baseline, the agent’s default mode could be to introduce societal considerations as essential context while defaulting control to the user, facilitating a more dynamic and transparent alignment. 

A second trade-off exists between transparency and usability. While making an agent’s reasoning transparent is a core tenet of trustworthiness, detailed explanations can create cognitive overhead making responses lengthy and difficult to parse. Research indicates that when faced with complex explanations, users may either ignore them or, paradoxically, develop an over-reliance on the system, as the mere presence of an explanation can create misplaced confidence \citep{Bansal2021-yw}. This places an undue cognitive burden of evaluating complex AI reasoning on the user, potentially undermining the goal of fostering critical engagement. This challenge calls for more dedicated research into effective design patterns for presenting information, such as interactive explanation interfaces, that allow users to drill down into evidence behind claims and validate agent reasoning if they want a justification of the material presented to them \citep{Zhou2025-gt}. 

Relatedly, the proposed emphasis on greater model transparency and process-based justification hinges on non-trivial advances in model interpretability. Current techniques like Chain-of-Thought (CoT) reasoning are often brittle and may act as post-hoc rationalizations rather than faithful traces of the model's actual inference process \citep{turpin2023languagemodelsdontsay, chen2025reasoningmodelsdontsay}. However, emerging research within the field of mechanistic interpretability, which aims to reverse-engineer the specific neural circuits responsible for complex behaviors, could one day allow auditors to verify the actual computational pathways an agent uses to weigh evidence or identify sources. Approaches like causal tracing enable researchers to pinpoint which pieces of information or internal representations were decisive in a conclusion by intervening in the model’s forward pass \cite[e.g.][]{meng_locating_2022, meng_mass-editing_2023, goldowsky-dill_localizing_2023, geva_dissecting_2023}. And data influence functions allow researchers to identify parts of the training data that are influential for a particular reasoning stream \citep{ruis_procedural_2025, grosse_studying_2023, koh_understanding_2020}. While still nascent, these advances hold promise for grounding an agent's claims in a genuine inferential process. However, it is not guaranteed that these methods will scale to the largest frontier models, or that they can otherwise successfully provide explanations of circuits that may be extremely complex or lacking analogy to human concepts.

Our proposed framework also highlights important tensions between standardization and paternalism: What is an optimal amount of cognitive "friction" for individuals? When are efforts to uphold these epistemic standards justified, and can AI systems be effectively trained to uphold them? The mixed responses to social media companies' effort to curb misinformation on their platforms illustrates the difficulty of balancing applying legal mandates to curb epistemic harms (e.g. fake news) without undermining epistemic values such as freedom of expression \citep{Aswad2019-bk}. The political economy of AI further complicates this dynamic. As the industry moves towards monetization models predicated on user retention \citep{Jin2025-qu}, developers must navigate the trade-off between building models that people actually want to use without impeding their long-term interests by optimizing for short-term metrics as a proxy for utility. Solving this coherently remains an open research challenge. While some users may resent AI that challenges them or makes them work harder \citep{Bucinca2021-ie}, others may actively seek agents aligned with their deeper epistemic goals. A potential path forward could be the development of a "marketplace of epistemic models," where users can select systems based on their preferred alignment philosophies. To function effectively, however, such a marketplace would require transparency through auditable standards, clear labeling, and independent benchmarking on critical metrics like epistemic competence and falsifiability. 

Finally, the technical infrastructure proposed—from verifiable agent credentials to cryptographic provenance chains—is contingent upon widespread adoption and addressing existing infrastructural gaps. If major developers or intermediaries were to opt out, it could create a fragmented and vulnerable ecosystem, rendering the entire accountability structure porous. Overcoming this collective action problem will require a concerted effort toward standardization, interoperability and regulation. While standards like C2PA are gaining traction among key stakeholders, they remain far from ubiquitous. This partial coverage creates second-order risks: the absence of provenance labels could be weaponized to cast doubt on legitimate content, and the infrastructure itself must be carefully architected to avoid enabling surveillance \citep{National-Institute-of-Standards-and-Technology-US-2024-qi,Chan2024-it}. Ultimately, however, these mechanisms mainly address transparency rather than veracity; they reveal \emph{who} generated content, but do not guarantee that the agent itself is trustworthy \footnote{A limitation mirrored in the application of cryptographic provenance to physical supply chains \citep{mvubu_blockchain_2024, kromes_fear_2024}.}. 

\section{Conclusion}
The potential emergence of autonomous epistemic AI agents heralds a transformation in the production, dissemination, and evaluation of knowledge. In this paper, we have argued that navigating this transformation safely and beneficially requires a holistic, socio-technical approach. We must concurrently align these agents with core human epistemic values, design them to be demonstrably trustworthy, and fortify the broader social and institutional ecosystem in which they will operate.
As our discussion of the limitations makes clear, this is no simple task. The path is fraught with profound ethical considerations and practical design trade-offs. Yet, these challenges do not negate the necessity of the framework; they reinforce its urgency. The alternative—socially un-moored epistemic systems—risks a further degradation of our already fragile collective episteme and missed opportunities in terms of augmenting and bettering it. The crucial work ahead is thus to deliberately and collaboratively architect the epistemic future we wish to inhabit.

\section*{Acknowledgements}
The authors would like to thank Michiel Bakker, Rishub Jain, Sebastien Krier, Chris Bregler, Seth Lazar, Michael Collins, Krisztian Balog, Petra Grutzik and Andrew Trask for their insightful feedback and discussions throughout the development of this work.

\bibliographystyle{abbrvnat}

\bibliography{references}

\end{document}